\documentclass{article}

\usepackage{arxiv}

\usepackage[utf8]{inputenc} 
\usepackage[T1]{fontenc}    
\usepackage{hyperref}       
\usepackage{url}            
\usepackage{booktabs}       
\usepackage{amsfonts}       
\usepackage{nicefrac}       
\usepackage{microtype}      
\usepackage{lipsum}
\usepackage{graphicx}
\graphicspath{ {./images/} }
\usepackage{url,hyperref,lineno,microtype,subcaption}
\usepackage[onehalfspacing]{setspace}
\usepackage{soul}
\usepackage{hyperref}
\usepackage{amsmath}
\usepackage{multirow}
\usepackage{multicol}
\usepackage{booktabs}

\title{Ensemble Learning of Myocardial Displacements for Myocardial Infarction Detection in Echocardiography}

\author{
 Tuan Nguyen$^{\dagger}$  \\
  VinUni-Illinois Smart Health Center, \\ VinUniversity, Hanoi, Vietnam
   \And
 Phi Nguyen$^{\dagger}$ \\
  VNU University of Engineering and Technology, \\ Hanoi, Vietnam 
  \And
 Dai Tran \\
  VNU University of Engineering and Technology \\ Hanoi, Vietnam 
  \And 
 Hung Pham  \\
  Vietnam National Heart Institute,  \\ Bach Mai Hospital, Hanoi, Vietnam
   \And
    Quang Nguyen  \\
 Vietnam National Heart Institute,  \\ Bach Mai Hospital, Hanoi, Vietnam
   \And
    Thanh Le  \\
  Vietnam National Heart Institute,  \\ Bach Mai Hospital, Hanoi, Vietnam
   \And
    Hanh Van  \\
  Vietnam National Heart Institute,  \\ Bach Mai Hospital, Hanoi, Vietnam
   \And
    Bach Do  \\
  Vietnam National Heart Institute,  \\ Bach Mai Hospital, Hanoi, Vietnam
   \And Phuong Tran  \\
   VNU University of Engineering and Technology, \\ Hanoi, Vietnam 
   \And Vinh Le  \\
   VNU University of Engineering and Technology, \\ Hanoi, Vietnam 
   \And
   Long Tran$^{*}$  \\
  VNU University of Engineering and Technology, \\ Hanoi, Vietnam  \\
  Correspondence: \texttt{tqlong@vnu.edu.vn}  
   \And
    Hieu Pham$^{*}$  \\
  College of Engineering \& Computer Science, \\ VinUni-Illinois Smart Health Center,  \\ VinUniversity, Hanoi, Vietnam \\ Correspondence: \texttt{hieu.ph@vinuni.edu.vn} \\ \\ 
  $\dagger$ These authors contributed equally to this work and share first authorship.
}
\usepackage{url,hyperref,lineno,microtype,subcaption}
\usepackage[onehalfspacing]{setspace}
\usepackage{soul}
\usepackage{hyperref}

\usepackage{multirow}
\usepackage{multicol}
\usepackage{booktabs}

\begin{document}
\onecolumn



\maketitle

\begin{abstract}

\textbf{Background:} Early detection and localization of myocardial infarction (MI) can reduce the severity of cardiac damage through timely treatment interventions. In recent years, deep learning techniques have shown promise for detecting MI in echocardiographic images. Existing attempts typically formulate this task as classification and rely on a single segmentation model to estimate myocardial segment displacements. However, there has been no examination of how segmentation accuracy affects MI classification performance and the potential benefits of using ensemble learning approaches. Our study investigates this relationship and introduces a robust method that combines features from multiple segmentation models to improve MI classification performance by leveraging ensemble learning.

\textbf{Materials and Methods:} Our method combines myocardial segment displacement features from multiple segmentation models, which are then input into a typical classifier to estimate the risk of MI. We validated the proposed approach on two datasets: the public HMC-QU dataset (109 echocardiograms) for training and validation, and an E-Hospital dataset (60 echocardiograms) from a local clinical site in Vietnam for independent testing. Model performance was evaluated based on accuracy, sensitivity, and specificity.

\textbf{Results:} The proposed approach demonstrated excellent performance in detecting MI. It achieved an F1 score of 0.942, corresponding to an accuracy of 91.4\%, a sensitivity of 94.1\%, and a specificity of 88.3\%. The results showed that the proposed approach outperformed the state-of-the-art feature-based method, which had a precision of 85.2\%, a specificity of 70.1\%, a sensitivity of 85.9\%, an accuracy of 85.5\%, and an accuracy of 80.2\% on the HMC-QU dataset. On the external validation set, the proposed model still performed well, with an F1 score of 0.8, an accuracy of 76.7\%, a sensitivity of 77.8\%, and a specificity of 75.0\%.

\textbf{Conclusions:} Our study demonstrated the ability to accurately predict MI in echocardiograms by combining information from several segmentation models. Further research is necessary to determine its potential use in clinical settings as a tool to assist cardiologists and technicians with objective assessments and reduce dependence on operator subjectivity. Our research codes are available on GitHub at \url{https://github.com/vinuni-vishc/mi-detection-echo}.

 
\end{abstract}

\section{Introduction}
A myocardial infarction (MI), which is also called a heart attack, happens when blood flow to part of the heart is cut off due to a clot~\cite{esc2012third} and severely damage the heart tissue. Most of the time, this happens because one or more of the coronary arteries, which bring blood to the heart, are blocked. 
MI is a serious and potentially life-threatening condition that is the leading cause of death worldwide, affecting 32.4 million people each year~\cite{benjamin2017heart}. In the US solely, about 4 million people visit the emergency room each year with heart symptoms~\cite{esc2012third}.
According to a study undertaken by the World Health Organization~\cite{thygesen2007universal}, cardiologists use multiple diagnostic indicators such as pathology outcomes, biochemical marker values, electrocardiography (ECG) findings, and other imaging modalities to diagnose patients with MI~\cite{esc2012third}. However, pathology can only detect dead cells in the heart muscle~\cite{thygesen2007universal}. ECG cannot distinguish between MI and myocardial ischemia symptoms~\cite{thygesen2018fourth}, and the specificity of biochemical marker values (cardiac enzymes) is quite low~\cite{stillman2011assessment}. Due to these limitations, none of these techniques are adequate for early MI detection. Therefore, the most valuable tool for early diagnosis is an imaging technique known as echocardiography, which is applicable for both clinical and research applications. Echocardiography (ECHO) is a pivotal tool for a safe and real-time functional assessment of the cardiovascular system \cite{lang2015recommendations}. It is based on ultrasonography, a noninvasive imaging technique that is incredibly valuable for monitoring and diagnosing patients who are exceedingly vulnerable. Moreover, echocardiography is fast, inexpensive, accessible, portable, and carries the lowest risk among imaging techniques~\cite{gottdiener2004american, chatzizisis2013echocardiographic}.

With the availability of the MI datasets on echocardiography, machine learning (ML) algorithms have been used to detect MI by extracting features from echocardiography~\cite{chen2020deep}. Although they showed promising results, previous studies have largely focused on using features from segmentation models, but the assumption that good segmentation equates to strong classification has yet to be fully substantiated. In addition, current methods are still limited by using only features from a single model, which is a common problem in MI classification and can lead to poor performance on unseen data~\cite{degerli2021early, hamila2022fully}. We conducted experiments to determine the relationship between strong segmentation models and precise MI classification. Our results showed that utilizing the predictions from multiple models through ensemble methods can better identify the patterns and features from echocardiography, resulting in more trustworthy and accurate predictions. Therefore, in this work, we propose a new approach to MI classification by incorporating multiple segmentation models and ensemble learning techniques. Our main contributions are summarized as follows:
\begin{itemize}
    \item Our experiment results showed that there is no strong correlation between good segmentation models and accurate MI classifiers. The finding indicates that highly accurate segmentation of the left ventricle (LV) is not a key condition for accurate MI detection.
    
    \item We proposed an ensemble method to combine multiple features produced by different LV wall segmentation models. The experimental results show that the proposed ensemble method consistently outperformed the state-of-the-art methods based on single models across all metrics on both the public and external test sets. This result suggests that ensemble learning is successful at complementing the features of multiple feature extractors in a way that a single feature model could not.

    \item The effectiveness of our method was tested on an external dataset obtained from a local clinical source. The results revealed a decline in the model's performance compared to public data test sets. Possible reasons for the decrease in performance are presented, along with an examination of the implications of these results. Recommendations are also given for enhancing the model's performance in a clinical environment.
    
    \item The proposed method showed a higher agreement score (Cohen's kappa value) than single-feature methods, regardless of whether they used different sets of features. This high level of agreement is an important advantage of the proposed approach as it suggests that the model predictions are subject to variation due to different feature extractors. The high Cohen's kappa value indicates that our proposed method is reliable and well-suited for use in the classification of MI.
\end{itemize}
\section{Related work}
\label{sec:related-work}
MI detection has been a focus of research in the field of medical imaging, with various techniques being proposed to detect abnormalities in LV wall motion~\cite{sudarshan2014automated, kiranyaz2020left, chen2020deep, hamila2022fully}. With the aim to reduce the cost and time of diagnosis, computer-assisted diagnostic techniques have been developed in recent decades that aim to automate the detection of MI by using image processing and ML techniques~\cite{giger2001computer,doi2006diagnostic}. Very first studies use active contour-based models, such as~\cite{mishra2003ga, landgren2013segmentation, dong2016left} the snake technique introduced by Kass \cite{kass1988snakes}, which uses an elastic curve to detect lines, boundaries, and edges in an image. However, these methods can be impractical or even impossible to use in cases where the LV wall is not visible due to low contrast or a portion of the wall is missing~\cite{yu2006towards}.  Other MI detection techniques include motion estimation methods~\cite{mondillo2011speckle}, which track the displacement of the LV wall, but the accuracy of these methods can depend on the performance of speckle tracking and can lead to unreliable results~\cite{dandel2009strain}.

Due to limitations in extracting features while using solely image processing techniques, recent studies shift to deep learning to extract hidden features from echocardiography images and detect MI. Neural networks such as U-Net \cite{ronneberger2015u} and U-Net++ \cite{zhou2019unet++} have been widely used for semantic segmentation. \cite{zhang2018fully, leclerc2019deep} presented a large dataset of 2D echocardiography images and proposed a U-Net-based model to accurate segmentation of LV wall. ~\cite{degerli2021early} utilized the accurate segmentation of LV wall to detect MI. While studies have shown that deep learning models can be used to detect MI, to the best of our knowledge, no work has explored the direct correlation between a strong segmentation of the LV wall and MI detection.

In addition to developing individual deep learning models, ensemble techniques, which combine the predictions of multiple models, have been explored as a way to improve performance and robustness in the analysis of cardiac functions. For example, \cite{narula2016machine} improved the performance of heart's morphological and functional assessments by using majority vote from three ML models: support vector machine, random forest, and deep learning. \cite{zhang2021ensemble} also increased the diagnostic performance of coronary heart disease screening by stacking ML models. While these studies have shown that ensemble techniques can improve the performance of medical problems, there is still no work that has explored the use of ensemble techniques for MI detection in echocardiography images.


Another major concern while using deep learning models for MI detection is the inconsistency of the results~\cite{wang2020inconsistent}. The performance of deep learning models is highly dependent on the quality of the training data, which can be difficult to obtain\cite{sarker2021deep}. In addition, the performance of deep learning models can be affected by the choice of the model's architecture and hyperparameters, which can be difficult to determine~\cite{schmidhuber2015deep}. To address these issues, we propose a strategy for combining features from different models in order to improve the diagnostic performance of MI detection. The proposed approach is based on the idea that by combining the strengths of various techniques, we can overcome their individual limitations and achieve more accurate and reliable results. We believe that this approach has the potential to significantly advance the field of MI detection and improve patient care.

The remainder of this paper is organized as follows. Section \ref{sec:matandmet} introduces the benchmark HMC-QU dataset, our in-house dataset from E-Hospital, and the framework for determining LV wall motion for MI identification. We will also describe in this section the experimental setting used in this study. Next, Section \ref{sec:results} presents the quantitative and qualitative evaluations of the proposed approach on the HMC-QU dataset and the E-Hospital dataset. We analyze the MI detection performance using various segmentation architectures and classifiers. Finally, Section \ref{sec:conclu} discusses the experimental results and concludes the paper with some potential research directions.


\section{Materials and methods}
\label{sec:matandmet}



\begin{figure*}[hbt!]
	\includegraphics[width=1\textwidth]{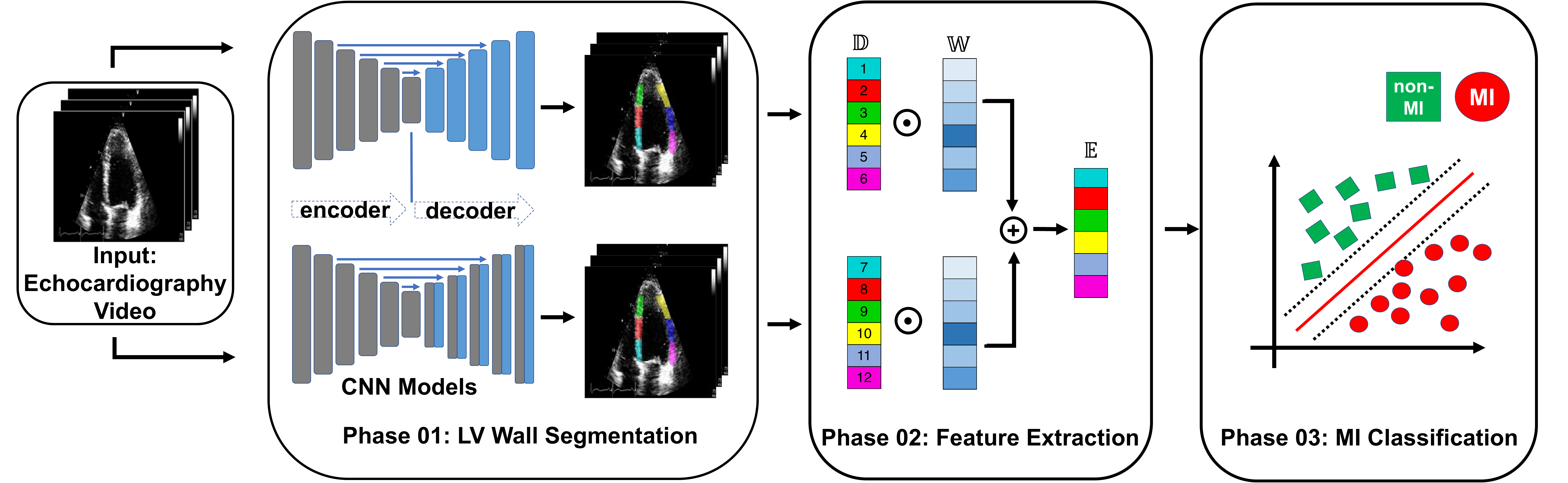} 
	\vspace{-1em}
        \caption{Overview of the proposed MI detection framework. Here $\mathbb{D}$ refers to the displacement of the heart muscle during a cardiac event, $\mathbb{W}$ refers to the weight assigned to different features within the ensemble model used for detection, and $\mathbb{E}$ refer to the ensemble of features used to identify MI.}
	\vspace{-1em}
	\label{fig:framework}
\end{figure*}

In this section, we introduce the proposed ensemble learning framework that addresses the challenge of MI detection. Figure \ref{fig:framework} illustrates  the  overall  architecture,  consisting  of three main phases: LV wall segmentation, feature engineering from myocardial segmentation displacement, and MI detection by traditional classification ML methods from ensemble models by weighting features from different segmentation models. Below we explain each phase in detail.

\begin{itemize}
    \item Phase 1 - LV wall segmentation: The goal of the first phase is to identify the myocardial boundary, which is a key indicator of heart function. Previous works often used only a single segmentation model to contour the LV. We further use multiple segmentation models and assess the accuracy of each segmentation model on each segment and on the whole LV.
    

    \item Phase 2 - Feature engineering from myocardial segmentation displacement: In this phase, we extract features from the segmented myocardial regions based on displacement over time. These features include measures such as strain, strain rate, and torsion, which are important indicators of myocardial function and can provide insight into the presence of MI. Because we used multiple segmentation models in Phase 1, we proposed a method to combine features from multiple segmentation results.

    \item Phase 3 - MI classification: In the final phase, traditional classification ML methods such as support vector machines, random forests, and logistic regression, are explored to detect MI from the extracted features. To further improve the performance of the proposed model, we also implement an ensemble learning approach by weighting the features from the different segmentation models. The weighting step takes into account the segmentation accuracy of each segmentation model. The performance of the proposed framework will be evaluated using common metrics, including sensitivity, specificity, precision, F1 score, and accuracy.
\end{itemize}



\subsection{DATASET}
\begin{figure}[hbt!]
	\includegraphics[width=1\columnwidth]{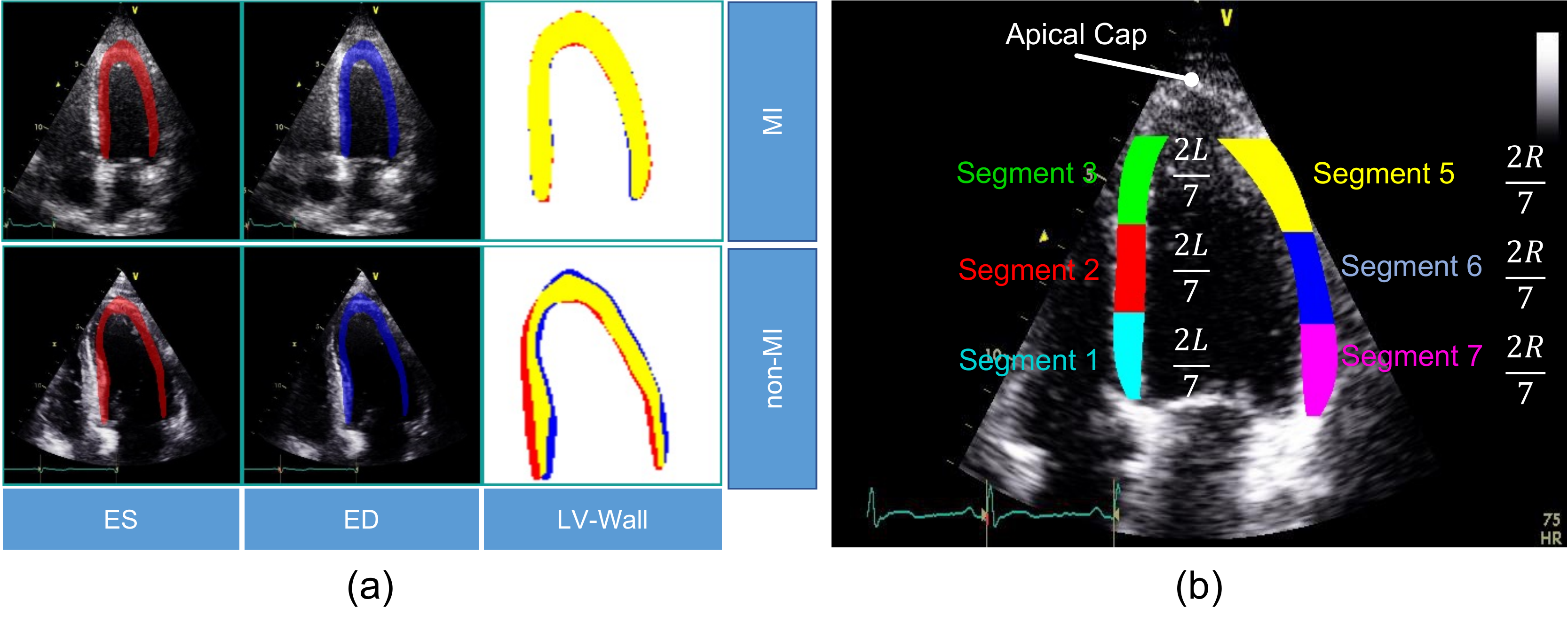} 
	\caption{(a) Segmentation mask of the LV wall for both an end-systolic frame and an end-diastolic frame, with one case representing a MI and the other case representing a non-MI case in the HCM-QU dataset. (b) Six segments of the LV wall that may be used to detect signs of a MI. The label ``L" represents the length from the bottom left corner to the apex of the LV, and the label ``R" represents the length from the bottom right corner to the apex of the LV.}
	\label{fig:dataset-and-segment}
\end{figure}

In this study, we used the public HMC-QU dataset \cite{degerli2021early} as the training and validation sets. The HMC-QU dataset 2D echocardiography recordings for the detection of MI and was established by cardiologists from the Hamad Medical Corporation, researchers from Qatar University and Tampere University. The ultrasound machines used to acquire the data were made by GE Healthcare, and the recordings have spatial and temporal resolutions that vary from 422$\times$636 to 768$\times$1024 and 25 frames per second, respectively. The collection includes 162 Apical Four Chambers (A4C) echos acquired between 2018 and 2019, but for the purposes of this study, a subset of 109 echos was used, resulting in a total of 2,349 images from 72 MI patients and 37 non-MI subjects. The remaining 53 echos were excluded because they did not include the entire LV wall, which was necessary for cardiologists to evaluate. As depicted in Figure \ref{fig:dataset-and-segment}a, the non-MI and MI cases have two frames: end-systolic and end-diastolic. The MI case has a significantly larger overlap region compared to the non-MI sample.
\begin{table*}[hbt!]
\begin{center}

\setlength{\tabcolsep}{0.5em} {
\renewcommand{\arraystretch}{1.2}
\caption {Sample counts of MI and non-MI patients by LV wall segments in HMC-QU and E-Hospital datasets.}

\begin{tabular}{lcccc}
\hline \hline



\multicolumn{1}{c}{\multirow{2}{*}{LV wall segments}} & \multicolumn{2}{c}{HMC-QU dataset}          & \multicolumn{2}{c}{E-Hospital dataset}      \\  \cmidrule(l){2-3} \cmidrule(l){4-5}
\multicolumn{1}{c}{}                                  & \# MI patients & \# non-MI patients & \# MI patients & \# non-MI patients \\

\hline
Segment-1        & 24                                      & 85                                     & 14                                      & 46                                     \\
Segment-2        & 43                                      & 66                                     & 15                                      & 45                                     \\
Segment-3        & 59                                      & 50                                     & 12                                      & 48                                     \\
Segment-5        & 44                                      & 65                                     & 7                                       & 53                                     \\
Segment-6        & 25                                      & 84                                     & 13                                      & 47                                     \\
Segment-7        & 15                                      & 94                                     & 15                                      & 45                                     \\ \hline
Patient-based    & 72                                      & 37                                     & 36                                      & 24                                    \\ \hline  \hline
\end{tabular}
\label{tab:tabledataset}
}
\end{center}
\end{table*}

To evaluate the effectiveness of our proposed method, we collected a dataset of patient records from E-Hospital, a local clinical site in Vietnam. The institutional review board at the clinical site approved the study, and we followed ethical guidelines in collecting and processing the data. We obtained 200 patient records from E-Hospital for this study, without specific selection criteria, covering the period from 2020 to 2021. The data were acquired using the Philip Affiniti 70G ultrasound imaging system, and we carefully examined all the data to ensure high-quality and visible anatomy structures. Only 60 data samples met our standards and were retained for further analysis. Three experienced doctors annotated the MI region in each sample, and another group of three doctors checked the annotations for accuracy and consistency. We used the annotated data for training and validation of our proposed model, which we stored in the Digital Imaging and Communications in Medicine (DICOM) format. The dataset includes only MI annotations and was used as a separate test set for evaluating the performance of our model.

\subsection{FEATURE ENGINEERING}
Similar to \cite{degerli2021early}, we extract the motion feature by performing two consecutive steps: LV (LV) segmentation and feature calculation. Below we explain these two steps in detail.

\textbf{LV segmentation.} We use a U-Net inspired convolutional network that consists of an encoding and down-sampling path, followed by a decoding and up-sampling path with skip links \cite{ronneberger2015u}. The model's output is a mask of the LV obtained through the Softmax function. During the training phase, we use cross-entropy as the loss function to optimize the model's parameters. Cross-entropy, which is defined as a measure of the difference between two probability distributions for a given random variable or set of events, is widely used for classification objectives. The binary cross-entropy is defined as
\begin{equation}
L_{BCE}(y, \hat{y})=-(y \log (\hat{y})+(1-y) \log (1-\hat{y})),
\end{equation}
where $\hat{y}$ is the predicted value by the prediction model.


\textbf{Feature calculation.} We analyze the segments of the LV wall to detect possible MI signatures. A standardized model~\cite{lang2015recommendations} recommends dividing the LV wall into seven segments, as depicted in Figure \ref{fig:dataset-and-segment}b. However, in our analysis, we only consider six of them, as the apical cap does not exhibit inward motion activity and should be skipped for the A4C view. For MI detection, we extract the displacement of the endocardial boundary points from the six segments. By evaluating the rate of displacement from the captured global motion of the LV wall, we aim to mimic the typical diagnosis of cardiologists, who assess segments that show a lack of motion as abnormal.

After segmenting the LV wall, we further extract its inner border to define the endocardial boundary, which is then divided into standardized six segments. The boundary segment displacements are calculated using the $L_1$ norm as follows
\begin{equation}
d_{L1} = |x^{t} - x^{t_{r}}| + |y^{t} - y^{t_{r}}|,
\label{eq1}
\end{equation}
where $x$ and $y$ are the pixel coordinates of the current frame $t$ and the reference frame $t_{r}$ (the first frame of one cycle). In order to capture the boundary segment motion more precisely, we take $N$ uniformly sampled pixels $p \in {(x_1, y_1), (x_2, y_2),..., (x_N, y_N)}$ on each frame $t$ for each segment $s$, and calculate the pair-wise distances $d_{s_t}$ between $t_{r}$ and $t$. The segment displacement for each frame is then calculated as 
\begin{equation}
d_{s_t} = \frac{1}{N}\sum_{n=1}^{N} |x_{n}^{t_r}-x_n^t|+|y_{n}^{t_r}-y_n^t|.
\label{eq2}
\end{equation}

Finally, we take the maximum pixel displacement of each segment, $d_s^{max} = max(d_{s_t})$, from the displacement curves, and normalize it to unity. The motion feature, $MF$, is then calculated as
\begin{equation}
MF={d_s^{max}}.
\label{eq3}
\end{equation}

\subsection{ENSEMBLE OF FEATURES}
\label{sec:ensemble}

Ensemble learning, which involves training a group of different classifiers and combining their predictions, has been shown to improve model performance and robustness \cite{rai2022hybrid}. During the process of feature engineering, we discovered that different segmentation models perform better on different segments (as shown in Table \ref{tab:segment_perf}). This meant that the performance of the classifier was limited by the performance of the segmentation model. To enable the classifier to benefit from multiple features, we calculated the accumulated vector by summing multiple feature vectors based on the accuracy of the segmentation models on the validation set. This approach has several benefits: 1) the weighting method does not require any training, so it can be used universally with different classifiers; 2) conventionally, stacking multiple feature vectors can increase the dimensionality of the accumulated feature vector, making it harder to train the classifier and limiting the number of features that can be used. In contrast, the weighted sum vector has a fixed dimension regardless of the number of segmentation models used.


Formally, given $n$ motion features vector $MF \in \mathbb{R}^6$ extracted from $n$ segmentation model with an objective metric for different segments $\mathbb{M} \in 	\mathbb{R}^6$. The weighted coefficient $\mathbb{W}$ is calculated as
\begin{equation}
\mathbb{W}_i =  \frac{\mathbb{M}_i}{\sum_{i=i}^{n}\mathbb{M}_i}.
\end{equation}

Accumulated feature vector $MF_{acc} \in \mathbb{R}^6$ is calculated as
\begin{equation}
MF_{acc} = \sum_{i=i}^{n} MF_i \odot \mathbb{W}_i.
\end{equation}

\subsection{MI DETECTION}
In the final step of the pipeline, we use ML to detect MI in an echocardiogram. To do this, we employ a variety of conventional supervised ML techniques, including 
support vector machine \cite{cortes1995support}, logistic regression \cite{cox1958regression}, decision trees \cite{breiman1996bagging}, and k-nearest neighbor \cite{cover1967nearest}. These techniques were chosen over more complex deep learning methods due to the small and imbalanced nature of our dataset, as well as the fact that the extracted features are more suited to simpler analysis. To fairly evaluate the performance of these classifiers, we use a stratified 5-fold cross-validation scheme. The details of their configuration, training, and testing are discussed in the following section.




\subsection{EVALUATION METRICS}
\label{sec:eval-metrics}
\textbf{LV Segmentation.} The Intersection over Union (IoU) metric, also known as the Jaccard index, is used to measure the overlap between the target mask and the prediction output. This metric is similar to the Dice coefficient, which is frequently used as a loss function during training. The IoU is calculated by dividing the intersection of the target and prediction by their union. It is written as

\begin{equation}
IoU=\frac{\text {target} \cap \text {prediction}}{\text {target} \cup \text {prediction}}.
\end{equation}

\textbf{MI Detection.} For the MI detection, we classify infarcted subjects as class-positive (MI) and normal, non-MI subjects as class-negative. In this case, the confusion matrix is formed as follows: TN is the number of correctly predicted non-MI subjects, TP is the number of correctly predicted MI patients, FN is the number of incorrectly detected MI patients as non-MI subjects, and FP is the number of incorrectly detected non-MI subjects as MI patients. The elements of the confusion matrix are calculated per video for MI detection. The standard performance evaluation metrics are defined as

\begin{equation}
\begin{aligned}
\mathrm{Sensitivity} &=\frac{\mathrm{TP}}{\mathrm{TP} + \mathrm{FN}},\\
\end{aligned}
\end{equation}
\begin{equation}
\begin{aligned}
\mathrm{Specificity} &=\frac{\mathrm{TN}}{\mathrm{TN} + \mathrm{FP}},\\
\end{aligned}
\end{equation}
\begin{equation}
\begin{aligned}
\mathrm{Accuracy} &=\frac{\mathrm{TP} + \mathrm{TN}}{\mathrm{TP} + \mathrm{TN} + \mathrm{FP} + \mathrm{FN} },\\
\end{aligned}
\end{equation}
\begin{equation}
\begin{aligned}
\mathrm{Precision} &=\frac{\mathrm{TP}}{\mathrm{TP} + \mathrm{FP}},\\
\end{aligned}
\end{equation}
\begin{equation}
\begin{aligned}
F(\beta)=\left(1+\beta^{2}\right) \frac{\text { Precision } \times \text { Sensitivity }}{\beta^{2} \times \text { Precision }+\text { Sensitivity }},
\end{aligned}
\end{equation}

where TP, FP, TN, and FN denote the numbers of true positive, false positive, true negative, and false negative cases, respectively. Sensitivity (also known as recall) is the ratio of correctly detected MI patients to all MI patients. Specificity is the ratio of correctly classified non-MI subjects to all non-MI subjects. Precision refers to the number of correctly detected MI patients over the total number of correctly detected samples. Accuracy is the ratio of correctly detected samples. F1 score is calculated as the harmonic average of precision and sensitivity, with a weighting of $\beta=1$ in the dataset.

\textbf{Prediction Reliability.} In addition to the performance of the prediction model, the reliability of the model is a crucial criterion for ML models used in medical applications. In this case, reliability can be interpreted as the consistency of the classifier's output regardless of the use of different feature extractors or combinations of feature extractors. We quantify this criterion by calculating the Cohen's Kappa~\cite{weighted_cohen_kappa} coefficient of the model's output based on the feature extractor. In this work, the Cohen's Kappa coefficient is calculated for three scenarios: 

\begin{itemize}
\item Agreement score between different predictors that use a single feature extractor. 
\item Agreement score between ensemble model using a different set of feature extractors.
\item Agreement score between cardiologist experts on single echo.
\end{itemize}

\subsection{EXPERIMENTAL SETTINGS}

\textbf{LV wall segemtation.} To evaluate the model, we employed a stratified 5-fold cross-validation scheme, in which we used 80\% of the available echos in the dataset to train the model and tested it on the remaining 20\% holdout (unseen) echos. During the training process, we trained all parameters in the network for 50 epochs using a learning rate of $1e-4$. The model was implemented in Pytorch and optimized using the Adam optimizer with parameters $\beta_1 = 0.9$ and $\beta_2 = 0.999$. The training and testing were conducted on a GTX 3090 GPU, and the number of total parameters ranged from 1 million to 23 million.

In our experiments, we explored a variety of architectures, including UNet \cite{ronneberger2015u}, UNet++ \cite{zhou2019unet++}, LinkNet \cite{chaurasia2017linknet}, DeepLabV3 \cite{chen2017rethinking}, and PAN \cite{li2018pyramid}, as well as several encoder architectures, including resnet18 \cite{he2016deep}, vgg11 \cite{simonyan2014very}, densenet121 \cite{huang2017densely}, efficientnet-b0 \cite{tan2019efficientnet}, and inceptionv4 \cite{szegedy2017inception}. The segmentation performance was evaluated on a pixel-level using the IoU score metric described in Section \ref{sec:eval-metrics}.

\textbf{MI detection.} 
In this experiment, we applied various supervised ML techniques for binary classification, including support vector machine (SVM), logistic regression (LR), decision trees (DT), and k-nearest neighbor (KNN). To begin, we collected a dataset consisting of input ensemble features and corresponding binary labels. Next, we split the dataset into a training set and a test set. The training set was used to train the different classification models, while the test set was used to evaluate their performance. 

For each of the models, we first fitted the model to the training data using default hyperparameter values. We then used the trained model to predict the labels for the instances in the test set. The performance of each model was evaluated using a variety of metrics, such as accuracy, precision, and recall.
Finally, we compared the performance of the different models to determine which one achieved the best results. For models used in our experiments, we used the default hyperparameters provided by the scikit-learn~\cite{scikit-learn} library, which is a widely used ML library in Python. This decision was made to ensure that our model could be easily reproduced and to facilitate comparisons with other studies that used similar methods.

\textbf{Model reliability analysis.} To evaluate the reliability of the predictions made by the different classification methods, we first randomly selected 20 examples from the test set. We then calculated the agreement score within each classification method (SVM, LR, DT, and KNN) that used only features derived from a single segmentation model. Each classification method was trained using four different sets of features. For the ensemble methods, we also calculated the agreement score between the four models, each of which was created using a different pair of features. Finally, we computed the human expert agreement score using the labels provided by three experts on the same echo samples. This allowed us to measure the similarity of the output of the feature extractors and classifiers on the same input.


\section{RESULTS}
\label{sec:results}
In this section, we present the results of our experiments on LV wall segmentation and MI classification using our proposed method on the HMC-QU validation test and a separate E-Hospital test set, respectively. We also evaluate the performance of our method through various metrics and techniques. First, we report the performance of our model on LV wall segmentation on the HMC-QU validation test on different LV wall segments, where we compare the state-of-the-art methods in the field. Next, we present the MI classification results on the HMC-QU and E-Hospital datasets and compare our model's performance against the baseline and other existing approaches. Finally, we discuss the reliability of our model by analyzing its agreement scores compared to existing methods and human experts.
\subsection{LV WALL SEGMENTATION ON HMC-QU VALIDATION TEST}
\begin{table*}[hbt!]
\begin{center}
\setlength{\tabcolsep}{0.5em}
\renewcommand{\arraystretch}{1.2}
\caption{Performance in IoU of different segmentation models on different segments of the LV wall on HMC-QU validation set.}
\vspace{1.0em}
\label{tab:segment_perf}

\begin{tabular}{lccccccl}
\hline \hline \multirow{2}{*}{Models} & \multicolumn{6}{c}{Segment} & \multirow{2}{*}{LV-Wall} \\  \cmidrule(l){2-7}
& 1  & 2  & 3  & 5  & 6  & 7  &  \\


\hline 
Unet++~\cite{zhou2019unet++}    & 0.851  & 0.905  & 0.873  & 0.823  & 0.873  &\textbf{ 0.932  } & 0.871\\
Unet~\cite{ronneberger2015u}      & 0.868  & 0.845  & 0.877  & \textbf{0.905}  &\textbf{ 0.979  }& 0.894 & \textbf{0.876}  \\
PAN~\cite{li2018pyramid}       & 0.858  &\textbf{ 0.938  }& 0.729  & 0.901  & 0.864  & 0.845 & 0.853  \\
LinkNet~\cite{chaurasia2017linknet}   &\textbf{ 0.895  }& 0.893  &\textbf{ 0.880  }& 0.868  & 0.809  & 0.859 & 0.867  \\
DeepLabV3~\cite{chen2017rethinking} & 0.870  & 0.766  & 0.854  & 0.883  & 0.939  & 0.839 & 0.860\\

\hline \hline
\end{tabular}
\end{center}
\end{table*}


Table \ref{tab:segment_perf} shows the LV wall segmentation averages (mean) results  for 5-fold CV  with different network architectures and encoders. The results indicate a stable IoU score between various types of architectures and encoders. The maximum IoU score can get is \textbf{0.876\%} when using UNet architecture and resnet18 encoder. Despite the fact that there is no significant variation in segmentation performance during validation, the segmentation models have different scores on different heart segments, as we have discussed in section \ref{sec:ensemble}.

\subsection{MI CLASSIFICATION ON TEST SET}

\begin{table*}[ht!]
\begin{center}
\setlength{\tabcolsep}{0.5em} {
\renewcommand{\arraystretch}{1.2}
\caption{MI accuracy based on segmentation features with different classifiers. The ensemble row shows the model that utilizes an ensemble of upper features on the HMC-QU dataset. The abbreviations `AE' and `WE' denote averaging ensemble and weighting ensemble, respectively.}
\vspace{1.0em}
\begin{tabular}{cccccccc}
    \hline \hline Classifier & Model & IoU & Sensitivity & Specificity & Precision & F1 score & ACC\\
    \hline 
     SVM & Single-PAN & 0.853 & 0.829 & 0.633 & 0.867 & 0.829 & 0.753\\
     SVM & Single-Unet & 0.871 & 0.764 & 0.857 & 0.914 & 0.831 & 0.797\\
     SVM & AE & - & 0.740 & 0.805 & 0.891 & 0.806 & 0.761 \\
     SVM & WE & - & \textbf{0.947} & 0.783 & 0.913 & \textbf{0.925} & \textbf{0.890}\\
    \hline  
     LR & Single-PAN & 0.853 & 0.767 & 0.783 & 0.902 & 0.816 & 0.756\\
     LR & Single-Unet & 0.871 & 0.769 & \textbf{0.960} & \textbf{0.967} & 0.851 & 0.820\\
     LR & AE & - & 0.771 & 0.776 & 0.858 & 0.806 & 0.774 \\
     LR & WE & - & 0.941 & 0.883 & 0.950 & \textbf{0.942} & \textbf{0.914}\\
    \hline  
     DT & Single-PAN & 0.853 & 0.787 & 0.760 & 0.907 & 0.836 & 0.784\\
     DT & Single-LinkNet & \textbf{0.867} & 0.773 & 0.607 & 0.809 & 0.788 & 0.722\\
     DT & AE & - & 0.726 & 0.795 & 0.887 & 0.794 & 0.749 \\
     DT & WE & - & 0.926 & 0.733 & 0.898 & \textbf{0.910} & \textbf{0.870}\\
     
    \hline  
     KNN & Single-PAN & 0.853 & 0.833 & 0.893 & 0.942 & 0.883 & 0.851\\
     KNN & Single-LinkNet & \textbf{0.867} & 0.812 & 0.783 & 0.908 & 0.844 & 0.787\\
     KNN & AE & - & 0.769 & 0.827 & 0.902 & 0.822 & 0.794 \\
     KNN & WE & - & 0.899 & 0.750 & 0.910 & \textbf{0.894} & \textbf{0.849}\\
     
    \hline \hline
\end{tabular}
\label{tab:exp-MI-HMC-QU}
}
\end{center}
\vspace{-1.3em}
\end{table*}

Table \ref{tab:exp-MI-HMC-QU} demonstrates that the performance of concatenated features from multiple segmentation models is superior compared to a single segmentation model and the performance across several classifications is consistent. Overall, the model achieved better performance by using ensemble features.

\begin{itemize}
\item Models with the same classifier performed similarly when using only one feature, on the other hand, ensemble models show better performance. Using SVM models, for instance, PAN or UNet segmentation features yielded comparable F1 scores of 0.829 for PAN and 0.831 for UNet. In contrast, the ensemble SVM model performed better with F1 score of 0.925. This improvement is rather uniform across classifiers by a substantial margin.

\item Upon comparing the performance of several classifiers, we found that there exist discrepancies in the approaches employed. Using an ensemble of two or three features, LR provided the best performance with F1 score of 0.942, while SVM, DT, and KNN only achieved F1 scores of 0.925, 0.910, and 0.894, respectively. When more features are added to the ensemble, however, this result becomes inconsistent, with LR's F1 score dropping to 0.905 and DT's F1 score reaching 0.931. This finding suggests that classifier selection should be based on experimentation with all classifiers.

\item Regarding the number of features used to construct the model ensemble. We have experimented with combining two and three features. We discovered that combining additional features does not produce better results. In reality, adding more features to a model might diminish its performance. In the case of the SVM classifier, adding three features decreased the F1 score from 0.925 to 0.903. This could indicate that adding more features to an ensemble could eventually increase the noise to the final features, eradicating the effectiveness of the advantage segment.

\item Our ensemble method showed significantly better performance compared to the average ensemble approach. our method achieved an F1 score of 0.925, while the average ensemble approach only achieved an F1 score of 0.806. This demonstrates the effectiveness of choosing the right regions from the LV wall, rather than average regions altogether.

\end{itemize}

\textbf{Correlation between good segmentation models and
accurate MI classifications.} In our experiment, we evaluated the performance of four different MI detection algorithms. Each algorithm used features from a weaker segmentation approach (PAN), and more accurate segmentation techniques (UNet and LinkNet). As shown in the table \ref{tab:exp-MI-HMC-QU}, While LinkNet and UNet achieved higher IoU scores of 0.871 and 0.867, respectively, algorithms that use PAN features still performs better than LinkNet and UNet with F1 score of 0.836 for DT-PAN and 0.788 DT-LinkNet. These results suggest that, while good segmentation is important for MI detection, it is not the only factor that determines the overall performance of the algorithm. In this case, the combination of advanced segmentation and effective feature extraction/classification techniques appears to be crucial for achieving optimal results.

\textbf{Comparison with previous state-of-the-art method.} Table \ref{tab:exp-sota} shows the performance of our ensemble models and the previous state-of-the-art method from \cite{degerli2021early}. In general, our ensemble LR model with two PAN features outperforms the state-of-the-art model presented in \cite{degerli2021early} by a significant margin. The F1 score for our model is 0.09 higher than the previous model, indicating that it is able to more accurately predict the outcome of interest. Additionally, when considering other evaluation metrics such as sensitivity, specificity, precision, and accuracy, our model consistently outperforms the model from \cite{degerli2021early}. This suggests that using a larger number of features can be beneficial for improving the performance of a ML model.

\begin{table*}[hbt!]
\begin{center}
\setlength{\tabcolsep}{0.5em} {
\renewcommand{\arraystretch}{1.2}
\caption{Comparison of performance metrics for our model and state-of-the-art method in a 5-fold cross-validation setting on HMC-QU dataset, both methods use the same experimental settings.}
\vspace{1.0em}

\begin{tabular}{lcccccc}
    \hline \hline Classifier & Sensitivity & Specificity & Precision & F1 score & ACC\\
    \hline LDA \cite{degerli2021early} & 0.785 & 0.701 & 0.838 & 0.806 & 0.756\\
    \hline RF \cite{degerli2021early} & 0.802 & 0.718 & 0.859 & 0.825 & 0.774\\
    
    \hline DT \cite{degerli2021early} & 0.790 & 0.586 & 0.804 & 0.794 & 0.726\\
    DT (\textbf{Ours}) & 0.926 & 0.733 & 0.898 & 0.910 & 0.870\\
    
    \hline SVM \cite{degerli2021early} & 0.859 & 0.701 & 0.855 & 0.852 & 0.802\\
    SVM (\textbf{Ours}) & \textbf{0.947} & \textbf{0.783} & \textbf{0.913} & \textbf{0.925} & \textbf{0.890}\\
    
    \hline \hline
\end{tabular}

    
    
    
\label{tab:exp-sota}
}
\end{center}
\vspace{-1.3em}
\end{table*}

\textbf{Performance of classification models on external test set.} Table \ref{tab:exp-MI-E-HOS} shows the result of our ensemble model in comparison with the single-feature model on the local clinical site, E-Hospital. The performance of the ensemble model on the external local clinical test set was found to be highly correlated with the results obtained on the public data test set. In terms of F1 score, the ensemble model achieved a score of 0.824 on the local clinical test set, which was significantly higher than the score from single-feature and average ensemble models. This demonstrates the consistant of our methods on different dataset. Additionally, the best ensemble model had a sensitivity, of 0.806 on the local clinical test set, which was again higher than single feature model and average ensemble model. Overall, these results suggest that the MI classification model is reliable and effective in identifying MI events in both familiar and novel datasets.

We also discovered that the external test on the local clinical dataset yielded lower scores compared to the public test set. Specifically, the model performed with F1 score of 0.824 on the local clinical dataset, while it achieved F1 score of 0.942 on the public test set. This suggests that the model may not generalize as well to the local clinical dataset, possibly due to differences in the distribution of the data or the specific characteristics of the patient population represented in the dataset. Solving this problem may be a task for future work. There are several potential approaches that could be pursued in order to address this issue, some possible approaches could include using different types of ML algorithms, collecting additional data, or fine-tuning the model parameters in order to improve performance. Ultimately, it will be important to carefully evaluate the strengths and limitations of different approaches in order to identify the most promising direction for future work.

\begin{table*}[hbt!]
\begin{center}
\setlength{\tabcolsep}{0.5em} {
\renewcommand{\arraystretch}{1.2}
\caption{MI accuracy based on segmentation features with different classifier, ensemble row indicate model using ensemble of upper features on E-Hospital dataset. The abbreviations `AE' and `WE' denote averaging ensemble and weighting ensemble, respectively.}
\label{tab:exp-MI-E-HOS}
\vspace{1.0em}
\begin{tabular}{cccccccc}
    \hline \hline Classifier & Model & Sensitivity & Specificity &     Precision & F1\-score & ACC\\
    \hline SVM & Single-PAN & 0.722 & 0.583 & 0.722 & 0.722 & 0.667 \\
     SVM & Single-Unet & 0.694 & 0.667 & 0.758 & 0.725 & 0.683 \\
     SVM & AE & 0.667 & 0.750 & 0.800 & 0.727 & 0.700 \\
     SVM & WE  & \textbf{0.806} & 0.625 & 0.763 & \textbf{0.784} & 0.733 \\
    \hline  LR & Single-PAN & 0.694 & 0.625 & 0.735 & 0.714 & 0.667 \\
     LR & Single-Unet & 0.611 & \textbf{0.792} & 0.815 & 0.698 & 0.683 \\
     LR & AE & 0.722 & 0.708 & 0.788 & 0.754 & 0.717 \\
     LR & WE  & 0.778 & 0.750 & 0.824 & \textbf{0.800} & \textbf{0.767} \\
    \hline  
    DT & Single-LinkNet & 0.722 & 0.667 & 0.765 & 0.743 & 0.700 \\
     DT & Single-PAN & 0.694 & 0.750 & \textbf{0.806} & 0.746 & 0.717 \\
     DT & AE & 0.722 & 0.708 & 0.788 & 0.754 & 0.717 \\
     DT & WE & \textbf{0.806} & 0.667 & 0.784 & \textbf{0.795} & 0.750 \\
    \hline  KNN & Single-LinkNet & 0.667 & 0.542 & 0.686 & 0.676 & 0.617 \\
     KNN & Single-PAN  & 0.722 & 0.583 & 0.722 & 0.722 & 0.667 \\
     KNN & AE & 0.667 & 0.708 & 0.774 & 0.716 & 0.683 \\
     KNN & WE & \textbf{0.806} & 0.625 & 0.763 & \textbf{0.784} & 0.733 \\
     
    \hline \hline
\end{tabular}

}
\end{center}
\vspace{-1.2em}
\end{table*}
\newpage
\subsection{MODEL RELIABILITY}

\begin{table}[hbt!]

    \centering
    \caption{Cohen's Kappa coefficient of different classifers, compared between model using feature from single model, using our ensemble method and human expert. Here higher scores indicate better agreement among classifiers.}
    \setlength{\tabcolsep}{0.5em}{
    \renewcommand{\arraystretch}{1.2}
    \vspace{1.0em}
    \label{tab:kappa}
        \begin{tabular}{ccc}
        
            \hline  \hline
 			 Architecture & Model &  Cohen's Kappa coefficient \\
            \hline  
 			 SVM & Single-Feature  & 0.74 \\
             SVM & Ensemble-Feature       & \textbf{0.81} \\
            \hline  
             LR & Single-Feature   & 0.73 \\
             LR & Ensemble-Feature       & \textbf{0.79} \\
            \hline  
 			 DT & Single-Feature  & 0.75 \\
             DT & Ensemble-Feature   & \textbf{0.82} \\
            \hline 
             Human expert &   & 0.96 \\

        \hline \hline
        \end{tabular}
    }
\end{table}
    
Table \ref{tab:kappa} shows the agreement scores between the single-feature model and ensemble-feature models for different classifiers and a human expert. A comparison between the single-feature model and ensemble-feature models shows that the prediction of ensemble-feature models is more consistent than that of the single-feature model. For example, the agreement score of the model's predictions by SVM single-feature models is 0.74, which is not so consistent with the human expert (0.96). The agreement score of ensemble models ranges from 0.79 to 0.82, the prediction by ensemble models is more similar to that of a human expert than that of a single-feature model, but still not consistent enough to replace a human expert. 

We observed no significant difference in consistency among single-model classifiers using SVM, LR, and DT. The same is true when we used ensembles, but we found that the DT ensemble had slightly better performance than the SVM and LR ensembles, as reflected in agreement scores of 0.82 for LR, and 0.81 and 0.79 for SVM and LR, respectively.

The above assessment show that ensemble models can be more accurate than a single model because it can capture a wider range of patterns and relationships in the data. In this case, it appears that the ensemble model has a higher agreement score than the single-feature model, which suggests that it is making more accurate predictions. This could also be due to the fact that the ensemble model is able to consider multiple features, rather than just one, which allows it to better capture the complexity of the data. Overall, the higher agreement score of the ensemble model indicates that it is a more effective model for MI classification problem.


\section{CONCLUSION}
\label{sec:conclu}
We proposed an accurate and robust deep learning-based approach for the MI detection on echocardiograms. We showed that accurate segmentation models are not fully correlated with accurate MI classifiers, indicating that highly accurate segmentation of the LV is not a key factor for building an accurate MI detection system. We then proposed an ensemble method for combining multiple features provided by different LV segmentation models, which outperformed both single models and state-of-the-art methods on two echocardiogram datasets. Compared to these existing approaches, the proposed method demonstrated significant improvement across all evaluation metrics. We further illustrated that the proposed method shows a higher agreement score (Cohen's kappa value) than single-feature methods, regardless of the features used. This high level of agreement suggests that our predictions are subject to less variation due to different feature extractors, making our method reliable and well-suited for use in the classification of MI.

Our work opens up several potential directions for further exploration. First, an end-to-end system for MI detection, rather than the current three-stage approach, should be developed. Second, further studies can be conducted to investigate the causes of performance drop on the external dataset and implement methods for addressing these issues. Third, it might be interesting to extend our training method into  the  pretraining  stage  to  produce better pre-trained models. Additionally, further exploration of the factors that contribute to the success of ensemble learning in MI detection could be beneficial for improving the performance of future models.

\section*{Conflict of Interest Statement}
TN, and HP were employed by VinUni-Illinois Smart Health Center, VinUniversity. PN, VL, TN, and LT were employed by VNU University of Engineering and Technology. DT was employed by E Hospital. HP, QN, TL, BD, and PT were  employed by Bach Mai Hospital. The authors declare that this research was conducted in the absence of any commercial or financial relationships that could be construed as a potential conflict of interest.

\section*{Author Contributions}
TN, LT and HP designed the research. TN and PN drafted the manuscript. TN, PN, LT, VL, TN  and HP revised the manuscript. DT, HP, QN, TL, BD, and PT collected and annotated the data. All authors have approved the final version.

\section*{Funding}
The research is supported by the Vingroup Innovation Foundation (VINIF) under project code VINIF.2019.DA02, and it is also supported by the VinUni-Illinois Smart Health Center at VinUniversity.



\section*{Data Availability Statement}
The HMC-QU dataset used in this study can be found and freely downloaded on Kaggle at \url{https://www.kaggle.com/datasets/aysendegerli/hmcqu-dataset}.
\section*{Ethics statement}
In the handling and use of medical data, we are committed to upholding the ethical principles of respect for persons, beneficence, and non-maleficence. We recognize that medical data is highly sensitive and personal, and we will take all necessary measures to protect the confidentiality and privacy of individuals. We will only use medical data for the purposes for which it was collected, and we will obtain informed consent from individuals before collecting or using their data. We will also ensure that the use of medical data is necessary and justified, and we will take steps to minimize any potential harm to individuals. Overall, our aim is to use medical data in a responsible and ethical manner that respects the rights and dignity of individuals.


\bibliographystyle{abbrv} 
\bibliography{references.bib}

\begin{thebibliography}{10}

\bibitem{esc2012third}
J.~J. Bax, H.~B. (Germany), C.~C. (Italy), V.~D. (France), C.~D. (UK), R.~F.
  (Belgium), C.~F.-B. (France), D.~H. (Israel), A.~H.~T. Netherlands), et~al.
\newblock Third universal definition of myocardial infarction.
\newblock {\em Journal of the American College of Cardiology},
  60(16):1581--1598, 2012.

\bibitem{benjamin2017heart}
E.~J. Benjamin, M.~J. Blaha, S.~E. Chiuve, M.~Cushman, S.~R. Das, R.~Deo, S.~D.
  De~Ferranti, J.~Floyd, M.~Fornage, C.~Gillespie, et~al.
\newblock Heart disease and stroke statistics—2017 update: A report from the
  american heart association.
\newblock {\em Circulation}, 135(10):e146--e603, 2017.

\bibitem{breiman1996bagging}
L.~Breiman.
\newblock Bagging predictors.
\newblock {\em Machine Learning}, 24(2):123--140, 1996.

\bibitem{chatzizisis2013echocardiographic}
Y.~S. Chatzizisis, V.~L. Murthy, and S.~D. Solomon.
\newblock Echocardiographic evaluation of coronary artery disease.
\newblock {\em Coronary Artery Disease}, 24(7):613--623, 2013.

\bibitem{chaurasia2017linknet}
C.~E. Chaurasia~Abhishek.
\newblock Linknet: Exploiting encoder representations for efficient semantic
  segmentation.
\newblock In {\em 2017 IEEE Visual Communications and Image Processing (VCIP)},
  pages 1--4. IEEE, 2017.

\bibitem{chen2020deep}
C.~Chen, C.~Qin, H.~Qiu, G.~Tarroni, J.~Duan, W.~Bai, and D.~Rueckert.
\newblock Deep learning for cardiac image segmentation: A review.
\newblock {\em Frontiers in Cardiovascular Medicine}, 7:25, 2020.

\bibitem{chen2017rethinking}
L.-C. Chen, G.~Papandreou, F.~Schroff, and H.~Adam.
\newblock Rethinking atrous convolution for semantic image segmentation.
\newblock {\em arXiv preprint arXiv:1706.05587}, 2017.

\bibitem{weighted_cohen_kappa}
J.~Cohen.
\newblock Weighted kappa: Nominal scale agreement provision for scaled
  disagreement or partial credit.
\newblock {\em Psychological Bulletin}, 70(4):213--220, 1968.

\bibitem{cortes1995support}
C.~Cortes and V.~Vapnik.
\newblock Support-vector networks.
\newblock {\em Machine Learning}, 20(3):273--297, 1995.

\bibitem{cover1967nearest}
T.~Cover and P.~Hart.
\newblock Nearest neighbor pattern classification.
\newblock {\em IEEE Transactions on Information Theory}, 13(1):21--27, 1967.

\bibitem{cox1958regression}
D.~R. Cox.
\newblock The regression analysis of binary sequences.
\newblock {\em Journal of the Royal Statistical Society: Series B
  (Methodological)}, 20(2):215--232, 1958.

\bibitem{dandel2009strain}
M.~Dandel, H.~Lehmkuhl, C.~Knosalla, N.~Suramelashvili, and R.~Hetzer.
\newblock Strain and strain rate imaging by echocardiography-basic concepts and
  clinical applicability.
\newblock {\em Current Cardiology Reviews}, 5(2):133--148, 2009.

\bibitem{degerli2021early}
A.~Degerli, M.~Zabihi, S.~Kiranyaz, T.~Hamid, R.~Mazhar, R.~Hamila, and
  M.~Gabbouj.
\newblock Early detection of myocardial infarction in low-quality
  echocardiography.
\newblock {\em IEEE Access}, 9:34442--34453, 2021.

\bibitem{doi2006diagnostic}
K.~Doi.
\newblock Diagnostic imaging over the last 50 years: research and development
  in medical imaging science and technology.
\newblock {\em Physics in Medicine \& Biology}, 51(13):R5, 2006.

\bibitem{dong2016left}
S.~Dong, G.~Luo, G.~Sun, K.~Wang, and H.~Zhang.
\newblock A left ventricular segmentation method on \text{3D} echocardiography
  using deep learning and snake.
\newblock In {\em 2016 Computing in Cardiology Conference (CinC)}, pages
  473--476. IEEE, 2016.

\bibitem{giger2001computer}
M.~L. Giger, N.~Karssemeijer, and S.~G.~A. III.
\newblock Computer-aided diagnosis in medical imaging.
\newblock {\em IEEE Transactions on Medical Imaging}, 20(12):1205--1208, 2001.

\bibitem{gottdiener2004american}
J.~S. Gottdiener, J.~Bednarz, R.~Devereux, J.~Gardin, A.~Klein, W.~J. Manning,
  A.~Morehead, D.~Kitzman, J.~Oh, M.~Quinones, et~al.
\newblock American society of echocardiography recommendations for use of
  echocardiography in clinical trials.
\newblock {\em Journal of the American Society of Echocardiography},
  17(10):1086--1119, 2004.

\bibitem{hamila2022fully}
O.~Hamila, S.~Ramanna, C.~J. Henry, S.~Kiranyaz, R.~Hamila, R.~Mazhar, and
  T.~Hamid.
\newblock Fully automated \text{2D} and \text{3D} convolutional neural networks
  pipeline for video segmentation and myocardial infarction detection in
  echocardiography.
\newblock {\em Multimedia Tools and Applications}, 81(26):37417--37439, 2022.

\bibitem{he2016deep}
K.~He, X.~Zhang, S.~Ren, and J.~Sun.
\newblock Deep residual learning for image recognition.
\newblock In {\em Proceedings of IEEE Computer Society Conference on Computer
  Vision and Pattern Recognition}, pages 770--778, 2016.

\bibitem{huang2017densely}
G.~Huang, Z.~Liu, L.~Van Der~Maaten, and K.~Q. Weinberger.
\newblock Densely connected convolutional networks.
\newblock In {\em Proceedings of IEEE Computer Society Conference on Computer
  Vision and Pattern Recognition}, pages 4700--4708, 2017.

\bibitem{kass1988snakes}
M.~Kass, A.~Witkin, and D.~Terzopoulos.
\newblock Snakes: Active contour models.
\newblock {\em International Journal of Computer Vision}, 1(4):321--331, 1988.

\bibitem{kiranyaz2020left}
S.~Kiranyaz, A.~Degerli, T.~Hamid, R.~Mazhar, R.~E.~F. Ahmed, R.~Abouhasera,
  M.~Zabihi, J.~Malik, R.~Hamila, and M.~Gabbouj.
\newblock Left ventricular wall motion estimation by active polynomials for
  acute myocardial infarction detection.
\newblock {\em IEEE Access}, 8:210301--210317, 2020.

\bibitem{landgren2013segmentation}
M.~Landgren, N.~C. Overgaard, and A.~Heyden.
\newblock Segmentation of the left heart ventricle in ultrasound images using a
  region based snake.
\newblock In {\em Medical Imaging 2013: Image Processing}, volume 8669, pages
  1132--1140. SPIE, 2013.

\bibitem{lang2015recommendations}
R.~M. Lang, L.~P. Badano, V.~Mor-Avi, J.~Afilalo, A.~Armstrong, L.~Ernande,
  F.~A. Flachskampf, E.~Foster, S.~A. Goldstein, T.~Kuznetsova, et~al.
\newblock Recommendations for cardiac chamber quantification by
  echocardiography in adults: an update from the \text{American Society of
  Echocardiography and the European Association of Cardiovascular Imaging}.
\newblock {\em European Heart Journal Cardiovascular Imaging}, 16(3):233--271,
  2015.

\bibitem{leclerc2019deep}
S.~Leclerc, E.~Smistad, J.~Pedrosa, A.~{\O}stvik, F.~Cervenansky, F.~Espinosa,
  T.~Espeland, E.~A.~R. Berg, P.-M. Jodoin, T.~Grenier, et~al.
\newblock \text{Deep Learning for Segmentation Using an Open Large-Scale
  Dataset in 2D Echocardiography}.
\newblock {\em IEEE Transactions on Medical Imaging}, 38(9):2198--2210, 2019.

\bibitem{li2018pyramid}
H.~Li, P.~Xiong, J.~An, and L.~Wang.
\newblock \text{Pyramid Attention Network for Semantic Segmentation}.
\newblock {\em arXiv preprint arXiv:1805.10180}, 2018.

\bibitem{mishra2003ga}
A.~Mishra, P.~Dutta, and M.~Ghosh.
\newblock A ga based approach for boundary detection of left ventricle with
  echocardiographic image sequences.
\newblock {\em Image and Vision Computing}, 21(11):967--976, 2003.

\bibitem{mondillo2011speckle}
S.~Mondillo, M.~Galderisi, D.~Mele, M.~Cameli, V.~S. Lomoriello, V.~Zac{\`a},
  P.~Ballo, A.~D'Andrea, D.~Muraru, M.~Losi, et~al.
\newblock Speckle-tracking echocardiography: a new technique for assessing
  myocardial function.
\newblock {\em Journal of Ultrasound in Medicine}, 30(1):71--83, 2011.

\bibitem{narula2016machine}
S.~Narula, K.~Shameer, A.~M. Salem~Omar, J.~T. Dudley, and P.~P. Sengupta.
\newblock Machine-learning algorithms to automate morphological and functional
  assessments in \text{2D} echocardiography.
\newblock {\em Journal of the American College of Cardiology},
  68(21):2287--2295, 2016.

\bibitem{scikit-learn}
F.~Pedregosa, G.~Varoquaux, A.~Gramfort, V.~Michel, B.~Thirion, O.~Grisel,
  M.~Blondel, P.~Prettenhofer, R.~Weiss, V.~Dubourg, J.~Vanderplas, A.~Passos,
  D.~Cournapeau, M.~Brucher, M.~Perrot, and E.~Duchesnay.
\newblock Scikit-learn: Machine learning in {P}ython.
\newblock {\em Journal of Machine Learning Research}, 12:2825--2830, 2011.

\bibitem{rai2022hybrid}
H.~M. Rai and K.~Chatterjee.
\newblock Hybrid \text{CNN-LSTM} deep learning model and ensemble technique for
  automatic detection of myocardial infarction using big \text{ECG} data.
\newblock {\em Applied Intelligence}, 52(5):5366--5384, 2022.

\bibitem{ronneberger2015u}
O.~Ronneberger, P.~Fischer, and T.~Brox.
\newblock \text{U-Net:} convolutional networks for biomedical image
  segmentation.
\newblock In {\em International Conference on Medical Image Computing and
  Computer Assisted Intervention}, pages 234--241. Springer, 2015.

\bibitem{sarker2021deep}
I.~H. Sarker.
\newblock Deep learning: a comprehensive overview on techniques, taxonomy,
  applications and research directions.
\newblock {\em SN Computer Science}, 2(6):420, 2021.

\bibitem{schmidhuber2015deep}
J.~Schmidhuber.
\newblock \text{Deep Learning in Neural Networks: An Overview}.
\newblock {\em Neural Networks}, 61:85--117, 2015.

\bibitem{simonyan2014very}
K.~Simonyan and A.~Zisserman.
\newblock Very deep convolutional networks for large-scale image recognition.
\newblock {\em arXiv preprint arXiv:1409.1556}, 2014.

\bibitem{stillman2011assessment}
A.~E. Stillman, M.~Oudkerk, D.~Bluemke, J.~Bremerich, F.~P. Esteves, E.~V.
  Garcia, M.~Gutberlet, W.~G. Hundley, M.~Jerosch-Herold, D.~Kuijpers, et~al.
\newblock Assessment of acute myocardial infarction: current status and
  recommendations from the \text{North American} society for
  \text{Cardiovascular Imaging and the European Society of Cardiac Radiology}.
\newblock {\em The International Journal of Cardiovascular Imaging}, 27:7--24,
  2011.

\bibitem{sudarshan2014automated}
V.~Sudarshan, U.~R. Acharya, E.~Y.-K. Ng, C.~S. Meng, R.~San~Tan, and D.~N.
  Ghista.
\newblock Automated identification of infarcted myocardium tissue
  characterization using ultrasound images: a review.
\newblock {\em IEEE Reviews in Biomedical Engineering}, 8:86--97, 2014.

\bibitem{szegedy2017inception}
C.~Szegedy, S.~Ioffe, V.~Vanhoucke, and A.~Alemi.
\newblock Inception-v4, inception-resnet and the impact of residual connections
  on learning.
\newblock In {\em Proceedings of the AAAI conference on artificial
  intelligence}, volume~31, 2017.

\bibitem{tan2019efficientnet}
M.~Tan and Q.~Le.
\newblock Efficientnet: Rethinking model scaling for convolutional neural
  networks.
\newblock In {\em International Conference on Machine Learning}, pages
  6105--6114. PMLR, 2019.

\bibitem{thygesen2018fourth}
K.~Thygesen, J.~S. Alpert, A.~S. Jaffe, B.~R. Chaitman, J.~J. Bax, D.~A.
  Morrow, H.~D. White, and E.~G. on~behalf of the Joint European Society of
  Cardiology (ESC)/American College of Cardiology (ACC)/American Heart
  Association (AHA)/World Heart Federation (WHF) Task Force for the Universal
  Definition~of Myocardial~Infarction.
\newblock Fourth universal definition of myocardial infarction (2018).
\newblock {\em Circulation}, 138(20):e618--e651, 2018.

\bibitem{thygesen2007universal}
K.~Thygesen et~al.
\newblock Universal definition of myocardial infarction.
\newblock {\em Circulation}, 116(22):2634--2653, 2007.

\bibitem{wang2020inconsistent}
X.~Wang, G.~Liang, Y.~Zhang, H.~Blanton, Z.~Bessinger, and N.~Jacobs.
\newblock Inconsistent performance of deep learning models on mammogram
  classification.
\newblock {\em Journal of the American College of Radiology}, 17(6):796--803,
  2020.

\bibitem{yu2006towards}
W.~Yu, P.~Yan, A.~J. Sinusas, K.~Thiele, and J.~S. Duncan.
\newblock Towards pointwise motion tracking in echocardiographic image
  sequences--comparing the reliability of different features for speckle
  tracking.
\newblock {\em Medical Image Analysis}, 10(4):495--508, 2006.

\bibitem{zhang2018fully}
J.~Zhang, S.~Gajjala, P.~Agrawal, G.~H. Tison, L.~A. Hallock,
  L.~Beussink-Nelson, M.~H. Lassen, E.~Fan, M.~A. Aras, C.~Jordan, et~al.
\newblock Fully automated echocardiogram interpretation in clinical practice:
  feasibility and diagnostic accuracy.
\newblock {\em Circulation}, 138(16):1623--1635, 2018.

\bibitem{zhang2021ensemble}
J.~Zhang, H.~Zhu, Y.~Chen, C.~Yang, H.~Cheng, Y.~Li, W.~Zhong, and F.~Wang.
\newblock Ensemble machine learning approach for screening of coronary heart
  disease based on echocardiography and risk factors.
\newblock {\em BMC Medical Informatics and Decision Making}, 21(1):1--13, 2021.

\bibitem{zhou2019unet++}
Z.~Zhou, M.~M.~R. Siddiquee, N.~Tajbakhsh, and J.~Liang.
\newblock \text{UNet++:} redesigning skip connections to exploit multiscale
  features in image segmentation.
\newblock {\em IEEE Transactions on Medical Imaging}, 39(6):1856--1867, 2019.

\end{thebibliography}

\end{document}